\documentclass[letterpaper]{article} 
\usepackage[]{aaai23}  
\usepackage{times}  
\usepackage{helvet}  
\usepackage{courier}  
\usepackage[hyphens]{url}  
\usepackage{graphicx} 
\usepackage{natbib}  
\usepackage{caption} 
\usepackage[ruled,linesnumbered,noend]{algorithm2e}
\usepackage{multirow}
\usepackage{subcaption}
\usepackage{amsmath,amssymb,amsthm}
\usepackage{subfiles}
\usepackage{listings}
\usepackage{xcolor}
\usepackage{comment}
\usepackage{tikz}
\usepackage{adjustbox}

\usetikzlibrary{decorations.pathreplacing,positioning, arrows.meta}

\definecolor{mygreen}{rgb}{0,0.6,0}
\definecolor{mygray}{rgb}{0.5,0.5,0.5}
\definecolor{mymauve}{rgb}{0.58,0,0.82}
\definecolor{myred}{rgb}{0.9,0.2,0.2}

\lstdefinelanguage{pddl}
{
  sensitive=false,    
  morecomment=[l]{;}, 
  alsoletter={:,-},   
  morekeywords={
    define,domain,problem,not,and,or,when,imply,forall,exists,either,
    :domain,:extends,:requirements,:types,:objects,:constants,
    :constraints,:ordered-substasks,:subtasks,:tasks,
    :predicates,:action,:durative-action,:duration,:method,:durative-method,
    :htn,:parameters,:precondition,:condition,:effect,:functions,
    :fluents,:primary-effect,:side-effect,:init,:goal,assign
    :strips,:adl,:equality,:task,:typing,:conditional-effects,:metric,
    :negative-preconditions,:disjunctive-preconditions,
    :existential-preconditions,:universal-preconditions,:ordered-subtasks,:ordering
  },
  keywords=[2]{object,at,start,over,all,end,always,at-most-once,sometime-before,sometime,sometime-after,hold-during,hold-between,hold-after,within,minimize,maximize,total-time} 
  keywords=[3]{calib_direction,image_direction,instrument,satellite,mode}, 
  keywords=[4]{calibrate,turn_approx,turn_precise,take_image,turn_to,activate_instrument,point_to,take_video,method_stereo,do_observation_stereo, do_observation,decrease_overall_quality,method_observe}, 
  keywords=[5]{observable,calibrated,pointing,supports,power_on,power_avail,on_board,calib_target,have_image,
  image-quality,calib-time,turn-time} 
}

\newtheorem{definition}{Definition}
\newtheorem{example}{Example}

\newcommand{\at}{\text{\it at}}

\newcommand{\before}{\text{\it before}}
\newcommand{\after}{\text{\it after}}
\renewcommand{\between}{\text{\it between}}
\newcommand{\atstart}{\text{\it at start}}
\newcommand{\atend}{\text{\it at end}}

\newcommand{\overall}{\text{\it overall}}

\newcommand{\name}{\text{\it name}}
\newcommand{\pre}{\text{\it precond}}

\newcommand{\effect}{\text{\it effect}}
\newcommand{\add}{\text{\it effect}^{+}}
\newcommand{\del}{\text{\it effect}^{-}}
\newcommand{\duration}{\text{\it duration}}
\newcommand{\tstart}{\text{\it start}}
\newcommand{\tend}{\text{\it end}}
\newcommand{\tinv}{\text{\it inv}}
\newcommand{\task}{\text{\it task}}
\newcommand{\tn}{\text{\it tn}}

\nocopyright

\title{HDDL 2.1: Towards Defining a Formalism and a Semantics \\for Temporal HTN Planning}

\author {
    Damien Pellier,\textsuperscript{\rm 1}
    Alexandre Albore,\textsuperscript{\rm 2}
    Humbert Fiorino,\textsuperscript{\rm 1}
    Rafael Bailon-Ruiz \textsuperscript{\rm 2}
}
\affiliations {
    \textsuperscript{\rm 1}University of Grenoble Alpes, LIG, Grenoble, France\\
      \{damien.pellier, humbert.fiorno\}@imag.fr\\
    \textsuperscript{\rm 2}ONERA/DTIS,  University of Toulouse, France\\
    \{alexandre.albore, rafael.bailon\_ruiz\}@onera.fr
}

\begin{document}
\maketitle

\begin{abstract}
  Real world applications as in industry and robotics need modelling rich and diverse automated planning problems. Their resolution usually requires coordinated and concurrent action execution. In several cases, these problems are naturally decomposed in a hierarchical way and expressed by a Hierarchical Task Network (HTN) formalism.
  HDDL, a hierarchical extension of the Planning Domain Definition Language (PDDL), unlike PDDL 2.1 does not allow to represent planning problems with numerical and temporal constraints, which are essential for real world applications. We propose to fill the gap between HDDL and these operational needs and to extend HDDL by taking inspiration from  PDDL 2.1 in order to express numerical and temporal expressions. This paper opens discussions on the semantics and the syntax needed for a future HDDL 2.1 extension.
\end{abstract}

\section{Introduction}

Real world applications of Automated Planning, like in industry and robotics, require modelling rich and diverse scenarios.
Such  planning problems are often naturally decomposed in a hierarchical way, with compound tasks that refine in different ways their execution model.
These real world applications of planning use both numerical and temporal constraints to define the agents synchronisation on collaborative tasks, and sub-task decomposition. In fact, concurrency between actions, their duration, and agents coordination in HTN problems are needed to find solutions for nontrivial tasks in complex scenarios and require to make explicit the representation of time \citep{ghallabnautraverso2016}.

The Hierarchical Task Network (HTN) formalism \citep{erol94} is used to express a wide variety of planning problems in  real-world applications, e.g., in task allocation for robot fleets \citep{Milot21}, video games \citep{Menif14} or industrial contexts such as software deployment \citep{Georgievski17}. Over the last years, much progress has been made in the field of hierarchical planning \citep{bercher19}. Novel systems based on the traditional, search-based techniques have been introduced~\citep{Bit-Monnot:16,ramoul17,Shivashankar17,Bercher17,Holler19,holler20,Holler21}, but also new techniques like the translation to STRIPS/ADL~\citep{Alford09,Alford16,behnke2022}, or revisited approaches like the translation to propositional logic~\citep{behnke2018totsat,Behnke2019orderchaos,Schreiber2019SAT,Schreiber21,behnke2021}. Despite these advances, not all planning systems use the same formalism to represent hierarchical task decomposition, making it difficult to compare approaches and promote HTN planning techniques. 

An extension of PDDL (Planning Domain Description Language)~\citep{mcdermott98}, called HDDL (Hierarchical Planning Domain Description Language) \citep{holler20}, has been proposed to address this issue. HDDL is based on PDDL 2.1 \citep{fox03} and is the result of several discussions within the planning community \citep{Holler19b} to fill the need of a standard language for the first Hierarchical Planning track of International Planning Competitions (IPC) in 2020. However, it was decided that the first version of HDDL would not include any of the temporal or numerical features of PDDL due to efforts to develop the language and related tools. In this paper, we illustrate the challenge of defining the semantics for a temporal extension of HDDL to meet the needs of the planning community and planning applications. 

\looseness=-1
Our motivation is grounded on the compelling need to devise applications involving autonomous systems. 
We propose to extend HDDL, by including elements of PDDL 2.1 and ANML ({\it Action Notation Modeling Language}) \citep{smith08}, to express temporal and numerical constraints.
This is intended to initiate discussions within the HTN community on establishing a standard -- HDDL 2.1~-- aimed at filling the gaps between existing hierarchical-temporal planning approaches. To that end, we make this preliminary extension of HDDL an open source project with a public repository, where we propose a full syntax as well as a set of benchmarks based on this extension\footnote{\url{https://github.com/pellierd/HDDL2.1}} and a parser for it, as part of the PDDL4J\footnote{\url{https://github.com/pellierd/pddl4j}} library \cite{Pellier18}.

The rest of the paper is organised as follows. In Section 2 we define the basic concepts of the proposed extension. In Sections 3 and 4 we set down the semantics for Temporal HTN planning. We conclude on the central aspects of this planning paradigm, and on future work. 

\section{Lifted Temporal HTN planning}


Throughout this section, we will use common notations from first-order logic, which we assume to be known. In the lifted formalism of HDDL 2.1, we assume for the sake of simplicity that all logical formulas are over a {\it function-free} first-order logic language ${\cal{L}} = (V, C, P)$. ${\cal{L}}$ consists of sufficiently many {\it constant} $c \in C$ representing the {\it objects} in the real world, {\it variables} $x \in V$ and {\it predicates} $p \in P$. Predicates have parameters that are either variables or constants. The predicate arity is the number of predicate parameters. For instance, $p(x, c)$ is a 2-arity predicate. We can now define {\it formulas} in a function-free first-order logic: (i) a predicate is a formula ; (ii) if ${\phi}$ and ${\psi}$ are formulas, then $\neg \phi$, $\phi \vee \psi$ and $\phi \wedge \psi$ are formulas ; (iii) if $\phi$ is a formula and $x$ is a variable, then $\forall x \phi$ is a formula. We define $\exists x \phi$ as $\neg \forall x \neg \phi$, and $\phi \rightarrow \psi$ as $\neg \phi \vee \psi$. $\forall$ and $\exists$ are respectively the universal and the existential quantifier. Conceptually, grounding a formula consists in generating a set of variable-free i.e. {\it ground} formulas \cite{helmert} as follows: a variable $x$ in a quantifier-free formula $\phi$ is eliminated by replacing $\phi$ with $|C|$ copies, one for each $c \in C$, where $x$ is substituted with $c$ in the respective copy. This substitution is denoted by $\phi[x/c]$. Regarding quantified formulas, $\forall x \phi$ is replaced by $\bigwedge_{c \in C} \phi[x/c]$ and $\exists x \phi$ by $\bigvee_{c \in C} \phi[x/c]$. We refer the reader to the work of by \citet{behnke20,ramoul17} for further details on grounding implementation. Note that it is always possible to transform a formula in function-free first-order logic into a finite set of ground formulas in propositional logic.

A {\it state} $s$ is a set of ground predicates. For the sake of conciseness, we will also consider $s$ as a Herbrand interpretation that assigns $true$ to all ground predicates in $s$, and $false$ to all ground predicates not in $s$. From this, a truth value can be computed for every {\it ground} formula from ${\cal{L}}$ by using the usual rules for logical composition.
Without loss of generality, a formula (not necessarily ground) $\phi$ is true in $s$ if and only if grounding $\phi$ generates at least one ground formula true in $s$. We will use the notation $s \models \phi$ to mean that the formula $\phi$ is true in $s$.



 A key concept in HTN planning and a fortiori in temporal HTN planning is the concept of {\it task}. Each task is given by a name and a list of parameters. We distinguish two kinds of tasks: the  primitive tasks (also called actions), and the abstract tasks (or compound tasks).
 Primitive tasks are carried out by  durative actions in the sense of classical temporal planning \cite{fox03}, while abstract tasks can be refined by applying methods that define the decomposition of the task into subtasks. The purpose of abstract tasks is not to induce a state transition. Instead, they refer to a predefined mapping to one or more tasks that can refine the abstract task. For instance, in the task of serving a dinner, {\it deliver-dinner(?food-style, ?place)} is the compound task consisting in performing first the task of serving the starters, then the main course, etc. In that sense,  {\it deliver-dinner(?food-style, ?place)} can be refined in: $\langle$ {\it serve-starters(?food-style, ?place)},  {\it serve-main-course(?food-style, ?place)}, etc.$\rangle$ This mapping between tasks is achieved by a set of decomposition methods, namely the methods (Def. \ref{def:met}) and the Temporal Task Networks (Def. \ref{def:TNN}).


 We first define the planning domain and problem for Temporal HTN Planning.
 \begin{definition}
   A \emph{planning domain} $\cal{D}$ is a tuple $({\cal{L}}, {\cal{T}}, {\cal{J}}, \alpha, {\cal{A}}, {\cal{M}})$, where
   $\cal{L}$ is the first-order logic language,
   $\cal{T}$ is the set of tasks,
   $\cal{J}$ is the set of task identifiers\footnote{Task identifiers are arbitrary symbols, which serve as place holders for the actual tasks they represent. Identifiers are needed because tasks can occur multiple times within the same task network, as we will see below.},
   $\alpha: {\cal{J}} \rightarrow {\cal{T}}$ is the function that maps task identifiers to tasks,
   ${\cal{A}}$ is a set of actions constituted by \emph{snap} actions (Def. \ref{def:snap}) and \emph{durative} actions (Def. \ref{def:dur}),
   $\cal{M}$ is the set of methods (Def. \ref{def:met}). 
 \end{definition}

 The domain implicitly defines the set of all states $S$ defined over all subsets of all ground predicates in $\cal{L}$.

 \begin{definition}
 A \emph{planning problem} $\cal{P}$ is a tuple $({\cal{D}}, {s_0}, w_0, g)$, where $\cal{D}$ is a planning domain, $s_0 \in S$ is the initial state, $w_0$ is the initial temporal task network (not necessary ground), and $g$ is a formula (not necessary ground) describing the goal.
 \end{definition}


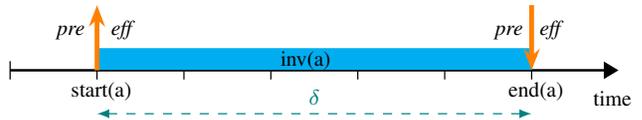
\begin{figure}
\begin{adjustbox}{width=\columnwidth}
\begin{tikzpicture}
\draw[thick, -Triangle] (0,0) -- (7cm,0) node[font=\scriptsize,below left=3pt and -8pt]{time};

\foreach \x in {0,1,...,6}
\draw (\x cm,3pt) -- (\x cm,-3pt);


\draw[cyan, line width=.25cm] (1.025 cm,0.13) -- +(4.975,0);
\draw [orange, arrows = {-Stealth[length=7pt]}, line width=.55mm] (1 cm, 0) -- +(0, 0.75);
\draw [orange, arrows = {-Stealth[length=7pt]}, line width=.55mm] (6 cm, 0.75) -- +(0, -0.75);


\draw[>=latex, <->, dashed, teal] (1,-.5) --  +(5,0);
\node[font=\scriptsize, text height=1.75ex,text depth=.5ex, teal] at (7/2,-.3) {$\delta$};

\node[font=\scriptsize, text height=1.75ex,text depth=.5ex] at (0.7,.5) {\emph{pre}};
\node[font=\scriptsize, text height=1.75ex,text depth=.5ex] at (1.3,.5) {\emph{eff}};
\node[font=\scriptsize, text height=1.75ex,text depth=.5ex] at (1.05,-.2) {start(a)};
\node[font=\scriptsize, text height=1.75ex,text depth=.5ex] at (5.74,.5) {\emph{pre}};
\node[font=\scriptsize, text height=1.75ex,text depth=.5ex] at (6.23,.5) {\emph{eff}};
\node[font=\scriptsize, text height=1.75ex,text depth=.5ex] at (6.05,-.2) {end(a)};
\node[font=\scriptsize, text height=1.75ex,text depth=.5ex] at (3.4,.15) {inv(a)};
\end{tikzpicture}
  \end{adjustbox}

\caption{Timeline of a durative action \emph{a} application.\label{fig:timeline} \vspace{-6mm}}
\end{figure}

 Let us start by defining the concepts of {\it snap} and {\it durative actions}, which are the primitive tasks, based on the definitions from \citet{abdulaziz22}. A snap action is an action whose execution is instantaneous in the sense of classical planning, meaning that it has a null duration 
 between checking the preconditions and applying the effects. 
 \begin{definition}[Snap Action]\label{def:snap} A \emph{snap action} $a$ is a tuple $\bigl(\name(a), \pre(a), \effect(a) \bigr)$, where $\name(a)$ is the name of $a$, the precondition $\pre(a)$ is a first-order formula, 
   and the effects $\effect(a) = \add(a) \cup \del(a)$ ($\add(a) \cap \del(a) = \emptyset$), $\add(a)$ and $\del(a)$ are conjunctions of predicates.

\end{definition}

\begin{definition}[Durative Action]\label{def:dur}
  A \emph{durative action} $a$ is a tuple $\bigl(\name(a), \tstart(a), \tend(a), \tinv(a), \delta\bigr)$: $\name(a)$ is the name of $a$; $\tstart(a)$ and $\tend(a)$ are snap actions; $\tinv(a)$ is a first-order formula that must hold in all the states after the execution of $start(a)$ and until the execution of $end(a)$, 
  and $\delta$ is the duration of $a$. 
\end{definition}
Actions do change the state of the world. Durative actions also change the time of the model, shifting it by a quantity $\delta$, as shown in Figure~\ref{fig:timeline}.
State transitions will be formally defined in Section \ref{THTN}.  

Unlike a primitive task, 
 a {\it compound task} does not directly  change the world state.
A compound task is identified by a name and defines the way other --possibly ordered-- tasks (either primitive or compound) must be achieved with respect to some constraints in order to refine it.
 Like primitive tasks, compound tasks have also a start event and an end event, which will be associated to dates in Section \ref{THTN}. 
In this sense, methods allow to refer tasks to temporal task networks.

\begin{definition}[Method]\label{def:met}
A \emph{method} $m$ is a tuple $\bigl(\name(m), \task(m), \tn(m)\bigr)$, where $\name(m)$ is the name of the method, $\task(m)$ is the task refined by the method, and $\tn(m)$ is the temporal task network decomposing $\task(m)$. 
\end{definition}

\begin{definition}[Temporal Task Network]\label{def:TNN}
A \emph{temporal task network} $w = ({\cal{I}}, {\cal{C}})$ is given by: 
\begin{itemize}

    \item $\cal{I} \subseteq \cal{J}$ is a set (possibly empty) set of tasks identifiers;

    \item $\cal{C}$ is the set of constraints, with ${\cal{C}} = < {\cal{C}}_o, {\cal{C}}_v, {\cal{C}}_d, {\cal{C}}_t >$:
      \begin{itemize}
      \item ${\cal{C}}_o$ is a set of temporal qualitative ordering constraints over the start or the end events of the tasks in $\cal{I}$. The possible qualitative temporal ordering are those from the classical point algebra \cite{broxvall03}: $<$, $\leq$, $>$, $\geq$, $=$ and $\neq$;

      \item ${\cal{C}}_v$ is a set of parameter constraints. Each constraint can bind two variables to be equal or non-equal, or similarly bind a variable to a constant; 

      \item ${\cal{C}}_d$ is a set of durative constraints over the duration of the tasks in $\cal{I}$;

      \item ${\cal{C}}_t$ is a set of temporal decomposition constraints of the form $(\at \ e \ \phi)$, expressing that some properties defined by the 
        formula $\phi$ must hold in the state at \mbox{date $e$.}
      \end{itemize}
\end{itemize}
The temporal task network $w$ implicitely defines a temporal ordered multi-set of tasks ${\cal{T}}' = \{ \alpha(i) \: | \: i \in {\cal{I}} \}$.
\end{definition}

A temporal task network explicits the decomposition of abstract tasks into subtasks. 
Note that a temporal task network is ground if all its variables are bound to constants, and primitive if all its tasks ${\cal{T}}'$ are primitive.

%
 \section{Temporal HTN Planning Semantics}
 \label{THTN}

 The solution of a temporal HTN planning problem is an {\it executable} temporal task network that is obtained from the problem initial task network by applying method decomposition and constraint satisfaction.

  Lifted problems are just a compact representation of their ground instances. Variable constraints are satisfied by the grounding, so there is no need to use them with ground instances. Therefore, for simplicity, this section defines the semantics of a lifted problem in terms of its ground instances. For details on the grounding process, the reader is referred to \cite{behnke20,ramoul17}.

Let us start by defining a temporal sequence of tasks.
\begin{definition}[Temporal Sequence of Tasks] 
  A \emph{temporal sequence of tasks} $\pi$ over a planning domain $\cal{D}$, is a sequence of tuples $\langle (t_0, e_0, \delta_0), \ldots,$ $(t_n, e_n, \delta_n) \rangle$ where $t_0, \ldots, t_n$ are ground tasks defined over $\cal{T}$, and for $0 \le i \le n$,  the natural numbers\footnote{
    Rational numbers are used in the definition from~\citet{fox03}. However, integers should be used for dates, because using rationals without adding further conditions can yield to an undecidable planning problem~\cite{barringer1986}.} %
  $e_i \in \mathbb{N}_{\geq 0}$ and $\delta_i \in \mathbb{N}_{\geq 0}$ are the starting date and the duration of the task $t_i$, respectively.
  For a temporal sequence of tasks $\pi$, the set of dates $\{e \ | \ (t, e, \delta) \in \pi\} \cup \{e + \delta \ | \ (t, e, \delta) \in \pi\}$ induces a sorted sequence $<e_0, \ldots, e_n>$ of \emph{happening events} of $\pi$. 
A temporal sequence of tasks  $\pi$ is primitive if and only if for every task  $(t, e, \delta) \in \pi$, $t$ is primitive, i.e. $t$ is carried out by an action (either snap or durative).
\end{definition}

The \emph{duration} of a task $t$ is generally unbounded, as the bound would be the sum of the durations of the tasks of which $t$ is compounded of. Only when a task $t$ is primitive, then $duration(t)$ is given by the duration $\delta$ of the action that achieves $t$.

In order to guarantee the exutability of concurrent plans, in the sense of the ``required concurrency'' as described by \citet{cushing2007temporal}, a central notion is {\it non-interference}, i.e. when preconditions and effects of snap actions do not overlap.
We consider that two snap actions $a$ and $b$ are \emph{not interfering} if and only if (i) $\pre(a) \cap \bigl(\add(b) \cup \del(b)\bigr) = \emptyset$, (ii) $\pre(b) \cap \bigl(\add(a) \cup \del(a)\bigr) = \emptyset$, (iii) $\add(a) \cap \del(b) = \emptyset$ and  $\add(b) \cap \del(a) = \emptyset$.

Two snap actions or more can be executed at the same time if they are pairwise non-interfering. The execution semantics of snap actions are similar to the semantics of $\forall$-step parallel plans \cite{rintanen06}, and used in PDDL~2.1 \cite{fox03}. With this notion in mind, we define an executable temporal sequence of tasks.

\begin{definition}[Executable Temporal Sequence of Tasks]
\label{def:exTST}
  A temporal sequence of tasks $\pi = \langle (t_0, e_0, \delta_0), \ldots,$ $(t_n, e_n, \delta_n) \rangle$ is executable in a state $s_0$ if and only if for every $(t, e, \delta)\! \in \!\pi$:
  (i) $t$ is primitive;
  (ii) $e$ is a happening event of $\pi$;
  (iii) 
  state $s_i$ at date $e_i$ transitions to a new state $s_{i+1}$ s.t.:
  given the set $B_{e_i}$ of snap actions being executed at $e_i$:
      $B_{e_i} = \bigl\{ \tstart(a_j) \,| \, (a_j, e_j, \delta_j) \in \pi$  and $e_j = e_i \}
      \cup  \{\tend(a_k)  | (a_k, e_k, \delta_k) \in \pi$ and $e_i + \delta_k = e_k\bigr\}$
     and given the set $I_{e_i}$ of the invariants holding at $e_i$:
      $I_{e_i} = \bigl\{ \tinv(a_j) \, | \, (a_j, e_j, \delta_j) \in \pi \wedge e_j < e_i < e_j + \delta_j\bigr\}$
    the transition of $s_i$ to  $s_{i+1}$, given that all the snap actions in $B_{e_i}$ are \emph{pairwise non-interfering}, is defined as   
       $s_i \models \pre(a) \text{ for every } a \in B_{e_i}$,
       $s_i \models \tinv(a)$  for every $\tinv(a) \in I_{e_i}$ and
       \mbox{$s_{i+1} = \left(s_i - \cup_{a \in B_{e_i}} \del(a)\right) \bigcup_{a \in B_{e_i}} \add(a)$.}
%
%
\end{definition}

Definition~\ref{def:exTST} can be extended to a temporal task network.

  \begin{definition}[Executable Temporal Task Network] A temporal task network $w = ({\cal{I}}, <\!{\cal{C}}_o, {\cal{C}}_v, {\cal{C}}_d, {\cal{C}}_t\!\!>)$ is executable in a state $s_0$ if and only if there exists an executable temporal sequence of tasks $\pi = \langle (\alpha(i_0), e_0, \delta_0), \ldots,$ $(\alpha(i_n), e_n, \delta_n)\rangle$  where $i_0, \ldots, i_n$ are task identifiers in $\cal{I}$ that matches the following conditions: (i) $\pi$ matches the temporal constraints ${\cal{C}}_o$, (ii) $\pi$ matches the duration constraints ${\cal{C}}_d$, and (iii) the sequence of states and their associated dates $\langle (s_0, e_0), \ldots,$ $(s_n, e_n) \rangle$ resulting from the execution of $\pi$ matches the constraints ${\cal{C}}_t$.
  \end{definition}

  It remains to define how to transform a temporal task network into another one by using method decomposition in order to obtain an executable task network,
  and what represents a temporal task network solution.
%


\begin{definition}[Decomposition]\label{def:decomp}
  %
  A task network $w_1= ({\cal{I}}^1, {\cal{C}}^{1})$ is decomposed into a new task network
  $w_2= ({\cal{I}}^2, {\cal{C}}^{2})$ by refinement by a method $m = \bigl(\name(m),$ $\task(m), ({\cal{I}}^m, {\cal{C}}^{m})\bigr)$
  if and only if there exists \mbox{$i \in {\cal{I}}^1$} such that $\alpha(i) = task(m)\;$ and
  $\;{\cal{I}}^2 = ({\cal{I}}^1 - \{i\}) \cup {\cal{I}}^m$, and
\begin{equation*} \label{eq1}
\begin{aligned}
{\cal{C}}^{2}_{o} &= \:{\cal{C}}^{1}_{o} \cup {\cal{C}}^{m}_{o} \\
& \cup \{ \tstart(t) \leq \tstart(t') \: | \: t= \min\left(e_{\tstart(\alpha(i))},e_{\tstart(\alpha(j))}\right),\\
&\quad t'= \max\left(e_{\tstart(\alpha(i))},e_{\tstart(\alpha(j))}\right), \: j \in {\cal{I}}^m \}  \\
& \cup \{ \tend(t) \geq \tend(t') \: | \: t=\max\left(e_{\tstart(\alpha(i))},e_{\tstart(\alpha(j))}\right),\\
& \quad t'= \min\left(e_{\tstart(\alpha(i))},e_{\tstart(\alpha(j))}\right), j \in {\cal{I}}^m \:  \}\\
\text{with } &e_{\tstart(\alpha(k))} \text{ the start date of } \alpha(k) \text{ for } k \in {\cal{I}}^1 \cup  {\cal{I}}^2.\\
{\cal{C}}^{2}_{v}  &=  {\cal{C}}^{1}_{v} \cup {\cal{C}}^{m}_{v},\;
{\cal{C}}^{2}_{d}  =  {\cal{C}}^{1}_{d} \cup {\cal{C}}^{m}_{d},\;
{\cal{C}}^{2}_{t}  =  {\cal{C}}^{1}_{t} \cup {\cal{C}}^{m}_{t}
\end{aligned}
\end{equation*}
\end{definition}

\begin{definition}[Temporal Task Network Solution]
  Let ${\cal{P}} = ({\cal{D}}, s_0, w_0, g)$ be a planning problem with $\cal{D} = ({\cal{L}}, {\cal{T}}, {\cal{A}}, {\cal{M}})$. A task network $w_s = ({\cal{I}}, {\cal{C}})$ is solution to a temporal HTN planning problem ${\cal{P}}$ \mbox{if and only if:}
  \begin{itemize}
    \item There exists a sequence of decompositions from $w_o$ to $w_s$ resulting from the application of the methods ${\cal{M}}$ of ${\cal{D}}$;
\item $w_s$ is executable and the temporal sequence of states resulting from its execution starts with $s_0$, and achieves a state $s \models g$.
\end{itemize}
\end{definition}

\section{Decomposition Constraint Semantics}
\label{Sec:Decomposition-Constraints-Semantics}

Decomposition constraints are conditions that must be satisfied by all the states visited while executing a solution task network. They are expressed through temporal modal operators over first-order formulas involving state predicates, as in PDDL. The semantics of the decomposition constraints can be formally specified similarly to the approach taken by \citet{gerevini05}. Let $w = ({\cal{I}}, {\cal{C}})$ be a ground task network, a state $s_0$ and a primitive temporal sequence of tasks $\pi = \langle (t_0, e_0, \delta_0), \ldots, (t_n, e_n, \delta_n) \rangle $ with $t_0 = \alpha(i_0), \ldots, t_n = \alpha(i_n)$ resulting from the decomposition of $w$.
We denote by $\tau = \langle (s_0, e_0), \ldots, (s_n, e_n) \rangle$ the  temporal sequence of states produced by the execution of $\pi$ in $s_0$, ordered according to its happening events, with \mbox{$i \leq 0 \leq n$}. Decomposition $w$ satisfies a constraint $(\at \ e \ \phi) \in {\cal{C}}_t$ iff $\exists (s_k, e_k) \in \tau$ such that $s_k \models \phi$. 
Note that it is required that every temporal constraint of the form $(\at \ e_i \ \phi) \in {\cal{C}}_t$ to be defined for $0 \leq i \leq n$ so as to avoid defining constraints that are out of the scope of the temporal task network $w$. Each decomposition constraint defined in HDDL~2.1 can be rewritten in terms of constraints of the form $(\at \ e\ \phi)$. The proposed HDDL extension distinguishes two types of decomposition constraints: (1)the \emph{temporal decomposition constraints} that define the constraints that must hold at specific happening events whose semantics is based on the plan trajectory constraints from PDDL~3.0 \cite{gerevini05} and (2) the \emph{classical decomposition constraints} ($\before$, $\after$, $\between$) used in HTN planning \citep{erol94} to represent constraints between tasks.

For the temporal decomposition constraints, we suggest to keep the same syntax and semantics as introduced in PDDL~3.0 to mantain a language consistency between different versions.
For classical decomposition constraints, it can be shown that they can be expressed in terms of the former, as are method precondition semantics ($\atstart, \atend, \overall$).

\section{Discussion and Conclusion}
\label{sec:conclusion}


We have introduced the main features of a HDDL version  for hierarchical temporal planning tasks.
This aims at bridging the gap between HTN planning and real world applications, where temporal features like concurrent actions, coordination, and hierarchical distribution of tasks, are prominent.
In our view, the absence of a unified language for temporal and numerical constraints in PDDL's evolution is an obstacle that  needs addressing. Although the community has developed various approaches to model complex planning problems, including temporal features and hierarchical task decomposition, the variety of language solutions hinders the development of common tools and solvers.

Action Notation Modeling Language (ANML)~\citep{smith08} has an expressivity close to what we seek here.
In ANML effects that happen at time intervals during an action duration can be specified. This  can also be represented in HDDL~2.1 by using constraint semantics and dividing actions with intermediate effects into separate durative actions.

Some planners propose benchmarks where ordering constraints are delayed, e.g., FAPE \cite{dvorak2014}. 
 For instance in ANML it is possible to specify that an action must happen at least some amount of time after the end of a previous action. With the proposed syntax and semantics, such expressivity can be reached by using auxiliary tasks that decompose in a durative primitive task (of the desired duration) with no effects. 

Time sampling represents another open question for this extension of HDDL with time.
Basically, two approaches exist. Sampling can be either constant ---when time is divided into regular-spaced discrete steps--- or with variable time steps instantiated when effects and preconditions are applied. The latter can benefit from the Simple Temporal Problem formalism to model the temporal features of the plan and to include timed initial effects, and Interval Temporal Logic can be used to define truth of formulas relative to time intervals, rather than time points \citep{BRESOLIN2014269}.


In order to be fully compatible with PDDL 3.0 features, the language HDDL 2.1 needs to include axioms and preferences, besides the associated parsing and validation tools. For this reason, the present work has to be seen as a baseline for the planning community to build upon.
In fact, many applications require features that have been (on purpose) omitted in this paper. Most importantly, we did not explicitly report a syntax, and we did not allow to define delays  with point algebra for ordering constraints. Other features are simply not detailed here, e.g., continuous effects.
Such language elements are left as natural extension of this paper for future work.



\end{document}